\documentclass[conference]{IEEEtran}
\usepackage{float}
\IEEEoverridecommandlockouts

\usepackage{cite}
\usepackage{amsmath,amssymb,amsfonts}
\usepackage{graphicx}
\usepackage{textcomp}
\usepackage{algpseudocode}
\usepackage{algorithm}
\usepackage{hyperref}
\usepackage{multirow}
\usepackage{booktabs}
\usepackage[table]{xcolor}
\usepackage{newfloat}
\usepackage{listings}
\def\BibTeX{{\rm B\kern-.05em{\sc i\kern-.025em b}\kern-.08em
    T\kern-.1667em\lower.7ex\hbox{E}\kern-.125emX}}
\floatstyle{ruled}
\newfloat{listing}{tb}{lst}{}
\floatname{listing}{Listing}

\begin{document}

\title{BlendFL: Blended Federated Learning for Handling Multimodal Data Heterogeneity  
}


\author{\IEEEauthorblockN{Alejandro Guerra-Manzanares$\dagger$,\thanks{$\dagger$ These authors contributed equally to this work.} Omar El-Herraoui$\dagger$, Michail Maniatakos and Farah E. Shamout}
\IEEEauthorblockA{Division of Engineering\\ New York University Abu Dhabi, Abu Dhabi, UAE\\
Email: \{alejandro.guerra, oe2015, 	michail.maniatakos, farah.shamout\}@nyu.edu}}


\maketitle

\begin{abstract}
One of the key challenges of collaborative machine learning, without data sharing, is multimodal data heterogeneity in real-world settings. While Federated Learning (FL) enables model training across multiple clients, existing frameworks, such as horizontal and vertical FL, are only effective in `ideal' settings that meet specific assumptions.  Hence, they struggle to address scenarios where neither all modalities nor all samples are represented across the participating clients. To address this gap, we propose BlendFL, a novel FL framework that seamlessly blends the principles of horizontal and vertical FL in a synchronized and non-restrictive fashion despite the asymmetry across clients. Specifically, any client within BlendFL can benefit from either of the approaches, or both simultaneously, according to its available dataset.  In addition, BlendFL features a decentralized inference mechanism, empowering clients to run collaboratively trained local models using available local data, thereby reducing latency and reliance on central servers for inference. We also introduce BlendAvg, an adaptive global model aggregation strategy that prioritizes collaborative model updates based on each client's performance. We trained and evaluated BlendFL and other state-of-the-art baselines on three classification tasks using a large-scale real-world multimodal medical dataset and a popular multimodal benchmark. Our results highlight BlendFL's superior performance for both multimodal and unimodal classification. Ablation studies demonstrate BlendFL's faster convergence compared to traditional approaches, accelerating collaborative learning. Overall, in our study we highlight the potential of BlendFL for handling multimodal data heterogeneity for collaborative learning in real-world settings where data privacy is crucial, such as in healthcare and finance. 

\end{abstract}


\section{Introduction}
\label{sec:introduction}
Healthcare institutions collect a variety of vast heterogeneous medical data \cite{yue2020deep}. The heterogeneity of the data, like in other domains, stems from the fact that each institution, also referred to as a client, collects different data modalities from a specific set of users that may or may not be represented at other clients \cite{milasheuski2024impact}. For example, one client may collect medical images while another collects laboratory test results. Hence, the overall aggregate dataset is considered to be multimodal, consisting of images, text, and/or numerical data, and heterogeneous due to varying levels of sparsity across the clients. Leveraging this aggregate dataset via collaborative learning to train centralized machine learning models could lead to improved performance, such as for enhanced diagnostics \cite{esteva2019guide}. However, in practice this is a very challenging task, not only due to privacy concerns \cite{rieke2020future}, but also due to the nature of the data.


In this scenario, Federated Learning (FL) can be used to collaboratively train machine learning models without sharing sensitive patient data. FL frameworks are particularly important in settings where data sharing is restricted due to privacy and security reasons, such as healthcare institutions or financial organizations \cite{rieke2020future}. However, traditional FL frameworks like Horizontal Federated Learning (HFL) and Vertical Federated Learning (VFL) face challenges when applied in real-world scenarios where data can be asymetrically fragmented and distributed unevenly across the clients \cite{wen2023survey}. HFL allows collaborative model training for clients that possess datasets with the same features but different data samples. Federated Averaging (FedAvg), proposed by \cite{mcmahan2017communication} is the most common form of HFL \cite{dayan2021federated}. In contrast, VFL deals with scenarios where clients hold different feature sets for the same data samples \cite{chen2020vafl}. While both frameworks may be effective in settings where clients are only allowed to participate if they meet specific conditions, they struggle to address hybrid or ill-defined scenarios where neither all features nor all samples are available across all the clients  \cite{yang2019federated}. This leads to suboptimal model training and inference capabilities, ultimately impeding non-conforming clients from participating in the federated network. 



To address this gap and advance the applicability of FL in complex, real-world environments, we introduce BlendFL — a novel framework that seamlessly integrates the full capabilities of HFL and VFL, addressing their inherent incompatibility in a unique framework. BlendFL is designed to handle different types of data fragmentation, enabling the training of collaborative models on both horizontally partitioned data, where distinct clients contribute different samples with a consistent feature set, and vertically partitioned data, where shared samples across clients are characterized by different feature sets. This dual capability allows clients to participate in the collaborative framework and benefit from HFL, VFL, or both, regardless of their share of features and samples. By accommodating varying data fragmentation distributions across clients (e.g., partial, paired, fragmented), BlendFL enhances model robustness and applicability in federated learning contexts, ensuring effective utilization of diverse data sources.

\begin{figure*}[ht]
\centerline{\includegraphics[width=0.8\linewidth]{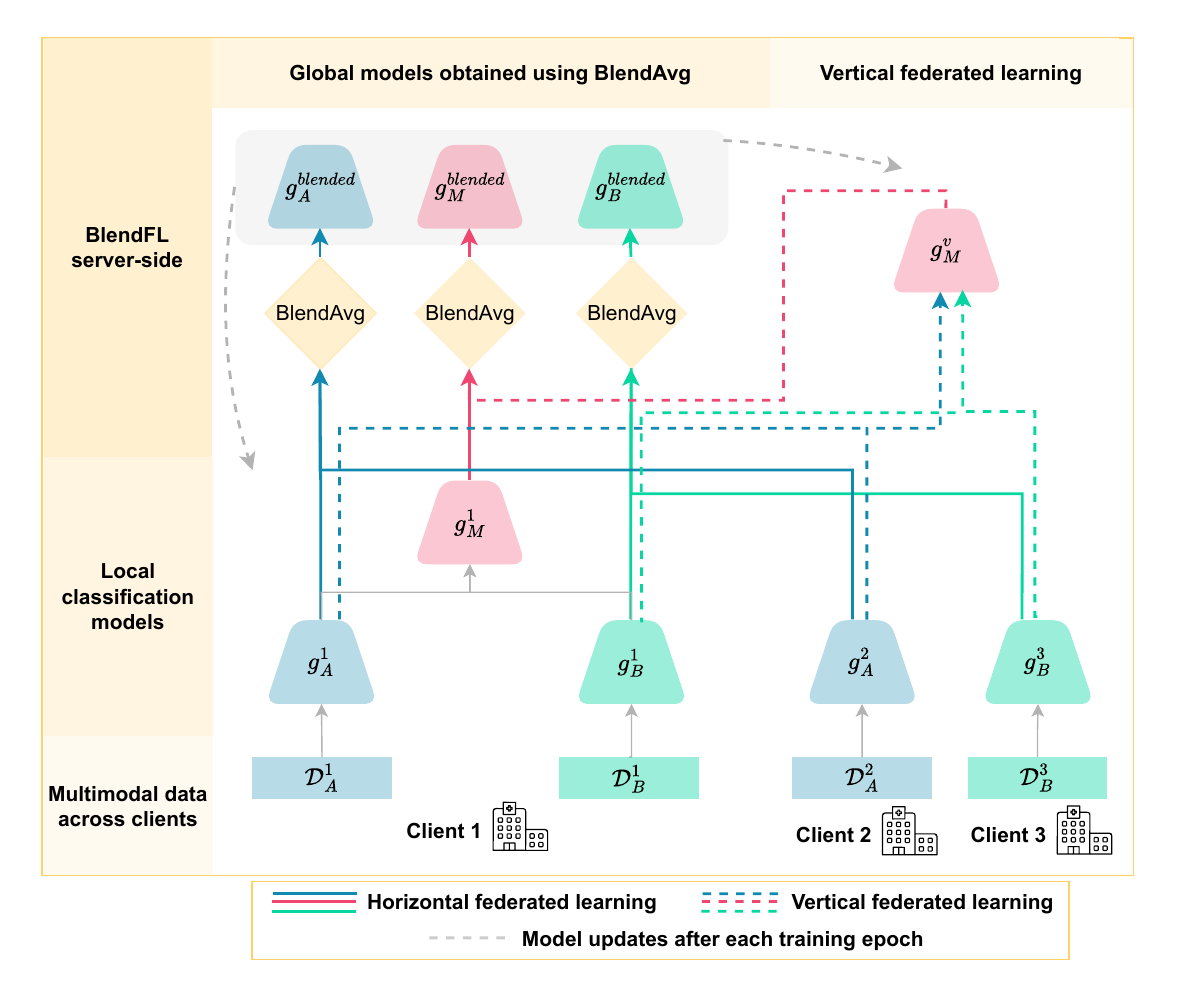}} \vspace{-3mm}
\caption{\small\textbf{BlendFL architecture}. We showcase the BlendFL architecture with 3 hospitals: 1 has multimodal data (client 1), while the other 2 hospitals have only unimodal data (clients 2 and 3). Each party $i$ has a specific dataset ($D^i_m$), for a specific modality $m$, where $m \in \{A, B\}$. Note that data can be fragmented (different modalities for the same patient are in different hospitals), partial (only one modality exists for a patient), or paired (both modalities are present in the same client). The dashed and dotted arrows indicate communication between models, either locally or between the hospitals and the server, as detailed in the legend. For clarity, we omitted \( f_A \) and \( f_B \), which are the feature extractors corresponding to each \( g_A \) and \( g_B \) classifier. Note that the arrows represent the data flow in one direction, to train the global models $g^{blended}_A$, $g^{blended}_B$, and $g^{blended}_M$. The grey dashed arrows represent the  distribution of the global models back to the clients after every epoch.}
\label{fig:architecture}
\end{figure*}


In summary, we make the following contributions:
\begin{itemize}
    \item We propose \textbf{BlendFL}, a novel FL framework that seamlessly integrates HFL and VFL for collaborative training on unevenly distributed, heterogeneous datasets. This integration overcomes the limitations of standard FL paradigms, allowing clients to join the collaborative effort and benefit from HFL, VFL, or both. An overview is shown in Figure~\ref{fig:architecture}.
    \item Unlike conventional VFL settings, BlendFL enables clients to perform independent multimodal and unimodal predictions after training, which we present as \textbf{decentralized inference}. This reduces dependency on a central server for inference tasks and minimizes communication overhead, which is especially beneficial in scenarios like healthcare where timely and local decision-making is critical.
    \item We introduce \textbf{BlendAvg}, a novel weighted parameter averaging scheme for model updates. Unlike traditional federated averaging, BlendAvg determines weights based on the performance of each local model on a representative validation set, rather than based on the volume of data each client has. This ensures that the best-performing models have a greater influence on the global model, promoting updates only if the validation performance improves, thereby preventing model degradation due to overfitting.
    \item We perform \textbf{extensive validation} of BlendFL on two datasets across three multimodal tasks, including a real-world clinical dataset that simulates a realistic scenario where multiple hospitals collaborate to predict patient outcomes without sharing private data, validating its practicality in a realistic setting. BlendFL outperforms seven state-of-the-art baselines across all tasks.
\end{itemize}

\section{Related Work}



Recent studies in FL have focused on either HFL or VFL \cite{banabilah2022federated}. In HFL, all clients hold datasets with identical features from different samples. One of the first HFL frameworks, FedAvg \cite{mcmahan2017communication}, utilizes a computation-then-aggregation strategy to merge local model updates from various clients to form a global model. Since then, a wide variety of enhancement strategies have been proposed, such as SCAFFOLD \cite{karimireddy2020scaffold}, which shares additional data with the server to improve model convergence, and FedMA \cite{wang2020federated}, which allows global and local models to be of different sizes. Note that all HFL frameworks assume that all client data have the same size and format \cite{zhang2020hybrid}.

In contrast, VFL deals with clients that hold different feature sets for common samples. Even though this setting has been notably less explored than HFL, one significant work is the FedBCD algorithm \cite{li2023effective}, which allows clients to perform multiple local updates before each communication, reducing communication overhead. Another study introduced FDML \cite{hu2019fdml}, which enables parties to perform asynchronous updates while mitigating the impact of stale information. Finally, SplitNN\cite{vepakomma2018split}, the most popular implementation of VFL, proposes the use of a cut layer that allows clients to train a partial network up to a specific layer before passing intermediate outputs to a server, enabling the completion of the training process without accessing raw data.


A third type of FL frameworks, the so-called hybrid approaches, attempt to combine HFL and VFL to address more complex data distributions across clients. \cite{zhang2020hybrid} propose the Hybrid Federated Matched Averaging (HyFEM) algorithm, which aligns features between local and global models using block coordinate descent, enhancing both privacy and model performance. Similarly, the FedHD \cite{gao2023fedhd} framework addresses the hybrid data challenge by incorporating gradient tracking and local stochastic gradient descent updates, which facilitates efficient communication and model training under partial client participation. In general, these hybrid methods are enhanced extensions of HFL incorporating some VFL capabilities, rather than frameworks that effectively combine both paradigms. To the best of our knowledge, BlendFL is the first framework to fully integrate HFL and VFL without imposing constraints or restrictions on clients. 

BlendFL introduces a unified framework which directly combines both major, and a priori incompatible, FL paradigms in a synchronized and coherent manner. Moreover, this seamless integration makes BlendFL proficient in handling complex, multimodal datasets, unlike other FL proposals, including hybrid frameworks, which focus exclusively on unimodal datasets (e.g., ModelNet40 \cite{zhang2020hybrid}, MNIST \cite{gao2023fedhd}). Specifically, BlendFL is designed to harness the full potential of multimodal data in a unique federated framework characterized by data fragmentation and client heterogeneity through a novel global model update strategy (BlendAvg). Furthermore, VFL frameworks do not support local inference \cite{zhang2023privacy}, requiring continued inter-client communication and synchronization for computing predictions. In contrast, BlendFL supports the development of robust unimodal and multimodal models that enable each client to perform local inference independently. 

\section{Methodology}

We first introduce the problem setting along with formal notation. For the sake of clarity, we define our problem setting within the healthcare domain. However, the framework is applicable to any domain where data privacy is important, and raw data sharing is not a feasible option for collaborative learning.

\subsection{Problem setting}

We assume that there are $N$ healthcare institutions, i.e., hospitals, $C=\{c^1, c^2, \cdots c^N \}$, that seek to collaboratively train a global model using heterogeneous data collected from a global set of $T$ patients, $U = \{u^1, u^2, \cdots, u^T\}$. The heterogeneity stems from the fact that patients have their data collected either at a single hospital or multiple hospitals, the data is multimodal (e.g., clinical imaging and electronic health records), and the hospitals are unable to share the raw private data of the clients directly with each other due to privacy reasons. For simplicity, we assume that each patient $u^i$ has either one or two data modalities collected across the hospitals, i.e. $x^i_A$ and/or $x^i_B$.

Each hospital has a local dataset  $\mathcal{D}^i$ collected from three possible types of patients: 

\begin{enumerate}
     \item Data of patients with a \textit{paired} set of modalities such that both $x_A$ and $x_B$ were collected at the same hospital, denoted as $\mathcal{D}_{paired (A,B)}$, 
    \item Data of patients with a \textit{fragmented} set of modalities, such that only one of the two modalities was collected at the hospital, while the other was collected at another hospital, denoted as $\mathcal{D}_{fragmented (A)}$ and $\mathcal{D}_{fragmented (B)}$, 
    \item Data of patients with a \textit{partial} set of modalities, such that only one modality was collected at the hospital, and the other modality was never collected otherwise, denoted as $\mathcal{D}_{partial (A)}$ and $\mathcal{D}_{partial(B)}$.
\end{enumerate}

Hence, a given hospital $c^i \in C$ has $\mathcal{D}^i= \{\mathcal{D}^i_A, \mathcal{D}^i_B\}$, where
\begin{equation}
    \mathcal{D}^i_A = \{\mathcal{D}^i_{paired (A)}, \mathcal{D}^i_{fragmented (A)}, \mathcal{D}^i_{partial (A)}\},
\end{equation}
\vspace{-2mm}
\begin{equation}
    \mathcal{D}^i_B = \{\mathcal{D}^i_{paired (B)}, \mathcal{D}^i_{fragmented (B)}, \mathcal{D}^i_{partial (B)}\},
\end{equation}

or, to reflect real-world scenarios, a combination of datasets collected from different kinds of patients, i.e. paired, fragmented and/or partial for specific data modalities $A$ and $B$.


Depending on the local set of data available $\mathcal{D}^i$, hospital $c^i$ also has a set of encoders and classifiers for the two modalities. For simplicity, we assume that all modality-specific encoders are uniform across hospitals, though the number of encoders each client has may vary depending on the data modalities available to them. The feature encoders, $f_A$ and $f_B$, process the modalities individually, such that:
\begin{equation}
\label{eq.3}
    h_A = f^i_A(x_A),  h_B = f^i_B(x_B),  
\end{equation}
where $h_A$ and $h_B$ are latent representations. The local unimodal and multimodal classifiers compute the predictions for a given task, $g_A$, $g_B$, and $g_M$, such that:
\begin{equation}
\label{eq.4}
    \hat{y}_A = g^i_A(h_A), \hat{y}_B = g^i_B(h_B), \hat{y}_M = g^i_M(h_A, h_B).
\end{equation}

Note that, in this setting, unimodal predictions are computed for local samples with missing modalities at a given hospital (i.e., fragmented and partial datasets), and multimodal predictions are computed for local multimodal samples (i.e. paired dataset).

\begin{algorithm}[t]
\caption{BlendFL Training Procedure}
\label{alg:BlendFL}
\begin{algorithmic}[1]
\Require $\mathcal{D}^k$, data partitions across clients $k \in \{1, \ldots, N\}$
\Require learning rate $\eta$, number of epochs $E$, number of clients $N$
\Ensure Trained models $g^{blended}_{A},\: g^{blended}_{B},\: g^{blended}_{M}$
\State Initialize server and client models $g^{k}_{A}, g^{k}_{B}, g^{k}_{M}, g^{v}_{M}$
\For{$e = 1$ to $E$}
    \For{each client $k$ in parallel}
        \State $x^k_A, y^k_A \gets$ Extract partial data from $\mathcal{D}_A$
        \State $x^k_B, y^k_B \gets$ Extract partial data from $\mathcal{D}_B$
        \State $g^k_{A} \gets \text{TrainLocalPartial}(x^k_{A}, y^k_{A})$
        \State $g^k_{B} \gets \text{TrainLocalPartial}(x^k_{B}, y^k_{B})$
    \EndFor
    \For{each client $k$ in parallel}
        \State $x^k_A, y^k_A \gets$ Extract fragmented data from $\mathcal{D}_A$
        \State $x^k_B, y^k_B \gets$ Extract fragmented data from $\mathcal{D}_B$
        \State $h^k_A \gets \text{ClientForwardPass}(x^k_{A}, y^k_{A})$
        \State $h^k_B \gets \text{ClientForwardPass}(x^k_{B}, y^k_{B})$
        \State $\text{SendFeaturesToServer}(h^k_A )$
        \State $\text{SendFeaturesToServer}(h^k_B )$
    \EndFor
    \State $\text{ServerAggregateFeatures}(h_A, h_B)$
    \State $\text{ServerForwardPass}()$
    \State $g^v_M \gets \text{ServerBackwardPass}()$
    \State $\text{ServerSendGradientsToClients}()$
    \For{each client $k$}
        \State $g^k_M \gets \text{ReceiveGradientsAndBackwardPass}()$
    \EndFor
    \For{each client $k$ in parallel}
        \If{client $k$ has local paired data}
            \State $x_A, x_B, y \gets$ Extract paired data from $\mathcal{D}^k$
            \State $g^k_{M} \gets \text{TrainLocalPaired}(x_A, x_B, y)$
        \EndIf
    \EndFor
    \State $\text{ClientsSendWeightsToAggregationServer}()$
    \State $g^{blended}_A,\: g^{blended}_B,\: g^{blended}_M \gets \text{ServerBlendAvg}()$
    \State $g^{k}_A,\: g^{k}_B,\: g^{k}_M \gets \text{LocalUpdate}(g^{blended}_A,\: g^{blended}_B,\: g^{blended}_M)$
\EndFor
\end{algorithmic}
\end{algorithm}


Next, we describe the local models (encoders and classifiers) at the hospitals, the global server and its components, and associated assumptions for the possible collaborative learning scenarios. The local models at the hospitals are as follows:

\begin{itemize}
    \item If the hospital only has paired samples (multimodal data for the same patients), then they locally have $f_A$, $f_B$, $g_A$, $g_B$, and $g_M$. 
    \item If a hospital only has fragmented and/or partial data for a given modality, then it would have $f_A$ and $g_A$, or $f_B$ and $g_B$, without loss of generality. 
\end{itemize}

Hospitals use their locally available encoders and classification models, both unimodal and multimodal, to perform local training with their available local datasets. Local encoder and classification models are used for local training and compute local predictions as described in Eq. \ref{eq.3} and Eq. \ref{eq.4}.

Then, if any two hospitals have complementary data based on overlap amongst patients (fragmented data, where the modalities for a patient are available but split across hospitals), they can collaborate, as in VFL, through the BlendFL server that has a global classifier trained using the complementary features of the local encoders:
\begin{equation}
\label{vertical}
    \hat{y}^v_{M} = g^v_M(h_{fragmented(A)}, h_{fragmented(B)}).
\end{equation}

Note that, for this step, we assume that all the collaborating hospitals share a common private database that includes identifiers for all individuals in the sample space, or that they implement a privacy-preserving dataset alignment technique, such as Private Set Intersection~\cite{morales2023private}, to match and pair data modalities for shared individuals prior to joint multimodal model training with $g_M^v$. 


Finally, to leverage the global set of data available at the hospitals, the BlendFL  server, as in HFL, collects the locally trained models from the hospitals and combines them to form global models (encoders and classifiers) by aggregating the weights of the local models after each training iteration, such that:
\vspace{-1mm}
\begin{gather}
\label{aggregation}
    g^{blended}_A = \text{BlendAvg}(g^i_A), \\
    g^{blended}_B = \text{BlendAvg}(g^i_B). 
\label{aggregation2}
\end{gather}

Note that this parameter aggregation step is performed using the BlendAvg strategy, described in the next section. The same procedure is applied for the multimodal models. However, in this case, both the locally trained multimodal models ($g^i_M$) and the collaborative trained model ($g^v_M$) are aggregated to obtain a blended classifier:
\begin{equation}
\label{blended}
    g^{blended}_M = \text{BlendAvg}(g^i_M, g^v_M),
\end{equation}

which is considered to be the final multimodal global model. Similarly, $g^{blended}_A$ and $g^{blended}_B$ are considered to be the final unimodal global models. 

After obtaining all global models, a training iteration is completed. Then, the server distributes all global models to the collaborating hospitals, which use them to update the weights of their respective local models. Specifically, $g^{blended}_M$, is used to update the weights of the global and local multimodal models, while $g^{blended}_A$ and $g^{blended}_B$ are used to update their parameters of all local unimodal models.

\subsection{BlendAvg}
\label{blendavg}

The Blended Averaging (BlendAvg) strategy aggregates the models' parameters based on each model's local improvement in terms of predictive power, as measured by performance metrics on a validation set. This predictive performance serves as a reliability metric for the local model parameters per hospital and is used to regulate the impact of local model contributions to the global model update. 

Consider the presence of $L$ models that are locally trained and used for global model averaging, where the $i$-th model has parameters indicated by $W_i$. In conventional HFL, the aggregation server uses the traditional FedAvg \cite{mcmahan2017communication} strategy to combine the parameters of the local models and output an averaged model, where all models contribute equally. On the other hand, BlendAvg proposes to aggregate the models' parameters proportional to the predictive power of each model. First, it calculates a weighting coefficient for each model based on a predictive performance score, denoted as $A_i$. For all local models, $A_i$ is calculated at the aggregation server using a private representative validation dataset that is randomly sampled from the collaborating clients. After evaluating the performance of each received model, the aggregation server proceeds with model aggregation as follows:

\begin{enumerate}
    \item \textbf{Measurement of local training improvement}. First, it calculates the improvement in predictive performance on the validation set for each model ($A_i$) compared to the previous global model performance on the same validation set ($A_{global}$). This assesses whether the local model training improved or not based on:
    \begin{equation}
        \Delta_i = A_i - A_{global},
    \end{equation}
    
    where $\Delta_i > 0$ indicates improvement by the local training on the validation set with respect to $A_{global}$, while $\Delta_i \leq 0$ indicates that the local training did not provide any improvement on the validation set with respect to $A_{global}$. Note that $A_{global}$ is calculated using the previous $g^{blended}_M$, on the same validation set, in the case of multimodal blending as defined in Eq.~\ref{blended}. For the unimodal encoders, $A_{global}$ is calculated using the previous $g^{blended}_A$ and $g^{blended}_B$ for Eq.~\ref{aggregation} and Eq.~\ref{aggregation2}, respectively.

    Note that this step and the subsequent steps are performed independently for each unimodal and multimodal models, as there is a global model for each one of them (see Figure \ref{fig:architecture}). However, for the sake of understanding and to avoid redundancy, the following steps are only described for a single generic computation (e.g., $g^{blended}_i$, where $i$ represents the $i$-th modality or multimodal data). 

    \item \textbf{Weight calculation and normalization}. 
    The subset of $l_d$ models, $l_d \in L$, reporting $\Delta_i \leq 0$ are discarded and not used for updating the global model. The subset of $l_u$ models, $l_u \in L$, with associated improvements ($\Delta_i > 0$) are considered for the global model update. Note that $l_d + l_u = L$. Next, the weighting coefficients ($\omega_i$) are calculated as follows: 
    \begin{equation}
        \omega_i =  \frac{\Delta_i}{\sum_{i=1}^{l_u} \Delta_i}.
    \end{equation}
    Each model in $l_u$ receives an associated weighting coefficient. Dividing by the sum of all improvements ensures that the sum of the weighting coefficients adds up to 1, assigning proportional weights to each model.

    \item \textbf{Weighted averaging}. The final model parameters ($W^{blended}_i$) are calculated as the weighted sum of each $l_u$ local model parameters ($W_i$) multiplied by its weighting coefficient $\omega_i$. Formally, it is defined as follows:
    \begin{equation}
        W^{blended}_i = \sum_{i=1}^{l_u} w_i \times W_i.
    \end{equation}

\end{enumerate}
Following this procedure, $W^{blended}_i$ only incorporates the parameters of the best performing models for the model update. This blending strategy ensures that each model's contribution to the final aggregated global model is proportional to its performance improvements, fostering a more adaptive and performance-oriented global model. Unlike traditional methods that average model parameters based on static criteria or data volume, like FedAvg, this approach allows for a dynamic adaptation to changes in model performance over time.
\begin{table*}[th]
\centering
\caption{Performance results of BlendFL, centralized learning, and federated baselines for clinical conditions prediction. Best results are shown in bold.}
\resizebox{\linewidth}{!}{\begin{tabular}{l>{\columncolor[RGB]{251,	218, 226}}c>{\columncolor[RGB]{251,	218, 226}}c>{\columncolor[RGB]{205,	246,	236}}c>{\columncolor[RGB]{205,	246,	236}}c>{\columncolor[RGB]{	255,	241,	209}}c>{\columncolor[RGB]{255,	241,	209}}c}
\toprule
\multirow{2}{*}{\textbf{Method}} & \multicolumn{2}{c}{\textbf{Multimodal}} & \multicolumn{2}{c}{\textbf{EHR}} & \multicolumn{2}{c}{\textbf{CXR}} \\
       & \textbf{AUROC} & \textbf{AUPRC} & \textbf{AUROC} & \textbf{AUPRC} & \textbf{AUROC} & \textbf{AUPRC} \\
\midrule
Centralized & 0.746 (0.733, 0.761) & 0.395 (0.357, 0.434) & 0.759 (0.742, 0.778) & 0.420 (0.380, 0.461) & 
0.707 (0.689, 0.725) & 0.408 (0.369, 0.448) \\
\midrule
FedAvg~\cite{mcmahan2017communication} & 0.713 (0.698, 0.729) & 0.358 (0.319, 0.398) & 0.748 (0.731, 0.766) & 0.403 (0.363, 0.445) & 0.677 (0.659, 0.696) & 0.400 (0.362, 0.440) \\
FedMA~\cite{wang2020federated} & 0.722 (0.707, 0.737) & 0.366 (0.327, 0.406) & 0.742 (0.725, 0.759) & 0.399 (0.360, 0.440) & 0.692 (0.674, 0.711) & 0.396 (0.357, 0.435) \\
FedProx~\cite{li2020federated} & 0.712 (0.697, 0.728) & 0.355 (0.316, 0.395) & 0.747 (0.730, 0.765) & 0.401 (0.361, 0.441) & 0.693 (0.674, 0.712) & 0.397 (0.358, 0.437) \\
FedNova~\cite{wang2020tackling} & 0.705 (0.690, 0.721) & 0.347 (0.307, 0.386) & 0.746 (0.729, 0.763) & 0.401 (0.361, 0.441) & 0.676 (0.657, 0.696) & 0.403 (0.364, 0.443) \\
One-Shot VFL~\cite{sun2023communication} & 0.711 (0.696, 0.726) & 0.352 (0.313, 0.392) & 0.742 (0.725, 0.760) & 0.391 (0.352, 0.432) & 0.681 (0.662, 0.701) & 0.402 (0.363, 0.442) \\
HFCL~\cite{elbir2022hybrid} & 0.698 (0.683, 0.714) & 0.341 (0.302, 0.381) & 0.734 (0.718, 0.752) & 0.382 (0.343, 0.422) & 0.684 (0.666, 0.703) & 0.388 (0.349, 0.427) \\
SplitNN~\cite{vepakomma2018split} & 0.706 (0.690, 0.722) & 0.341 (0.301, 0.381) & 0.741 (0.723, 0.760) & 0.391 (0.351, 0.432) & 0.680 (0.662, 0.699) & 0.398 (0.359, 0.437) \\
\rowcolor[gray]{0.9}BlendFL & \textbf{0.732 (0.717, 0.747)} & \textbf{0.375 (0.336, 0.415)} & \textbf{0.753 (0.735, 0.770)} & \textbf{0.408 (0.368, 0.448)} & \textbf{0.704 (0.686, 0.723)} & \textbf{0.430 (0.391, 0.469)} \\
\bottomrule
\end{tabular}}
\label{tab:pheno}
\end{table*}

\begin{table*}[th]
\centering
\caption{Performance results of BlendFL, centralized learning, and federated baselines for in-hospital mortality prediction. Best results are shown in bold.}
\resizebox{\linewidth}{!}{\begin{tabular}{l>{\columncolor[RGB]{251,	218, 226}}c>{\columncolor[RGB]{251,	218, 226}}c>{\columncolor[RGB]{205,	246,	236}}c>{\columncolor[RGB]{205,	246,	236}}c>{\columncolor[RGB]{	255,	241,	209}}c>{\columncolor[RGB]{255,	241,	209}}c}
\toprule
\multirow{2}{*}{\textbf{Method}} & \multicolumn{2}{c}{\textbf{Multimodal}} & \multicolumn{2}{c}{\textbf{EHR}} & \multicolumn{2}{c}{\textbf{CXR}} \\
       & \textbf{AUROC} & \textbf{AUPRC} & \textbf{AUROC} & \textbf{AUPRC} & \textbf{AUROC} & \textbf{AUPRC} \\
\midrule
Centralized & 0.866 (0.852, 0.880) & 0.501 (0.460, 0.541) & 0.867 (0.853, 0.881) & 0.491 (0.451, 0.531) & 0.729 (0.710, 0.748) & 0.181 (0.143, 0.220) \\
\midrule
FedAvg~\cite{mcmahan2017communication} & 0.844 (0.830, 0.858) & 0.432 (0.392, 0.472) & 0.856 (0.841, 0.871) & 0.498 (0.457, 0.538) & 0.688 (0.668, 0.708) & 0.167 (0.129, 0.205) \\
FedMA~\cite{wang2020federated} & 0.849 (0.835, 0.863) & \textbf{0.496 (0.456, 0.536)} & 0.851 (0.836, 0.866) & 0.508 (0.468, 0.548) & 0.630 (0.610, 0.650) & 0.141 (0.103, 0.179) \\
FedProx~\cite{li2020federated} & 0.845 (0.831, 0.859) & 0.484 (0.444, 0.524) & 0.849 (0.834, 0.864) & 0.514 (0.474, 0.554) & 0.606 (0.586, 0.626) & 0.127 (0.089, 0.165) \\
FedNova~\cite{wang2020tackling} & 0.843 (0.829, 0.857) & 0.477 (0.437, 0.517) & 0.849 (0.834, 0.864) & \textbf{0.515 (0.475, 0.555)} & 0.726 (0.708, 0.748) & 0.188 (0.150, 0.226) \\
One-Shot VFL~\cite{sun2023communication} & 0.848 (0.834, 0.862) & 0.495 (0.455, 0.535) & 0.854 (0.839, 0.869) & 0.507 (0.467, 0.547) & 0.717 (0.697, 0.737) & 0.189 (0.151, 0.227) \\
HFCL~\cite{elbir2022hybrid} & 0.839 (0.825, 0.853) & 0.471 (0.431, 0.511) & 0.844 (0.829, 0.859) & 0.506 (0.466, 0.546) & 0.582 (0.562, 0.602) & 0.125 (0.087, 0.163) \\
SplitNN~\cite{vepakomma2018split}  & 0.825 (0.810, 0.840) & 0.415 (0.375, 0.455) & 0.849 (0.834, 0.864) & 0.461 (0.421, 0.501) & 0.703 (0.683, 0.723) & 0.169 (0.131, 0.207) \\
\rowcolor[gray]{0.9}BlendFL    & \textbf{0.865 (0.851, 0.879)} & 0.494 (0.453, 0.534) & \textbf{0.864 (0.849, 0.879)} & 0.513 (0.472, 0.553) & \textbf{0.727 (0.708, 0.746)} & \textbf{0.195 (0.157, 0.233)} \\
\bottomrule
\end{tabular}}
\label{tab:mortality}
\end{table*}

\subsection{Execution details}

We depict an example of the BlendFL framework in Fig. \ref{alg:BlendFL}, which illustrates a scenario where three healthcare institutions participate in the collaborative training effort. In this scenario, each hospital has a subset of patients (samples). Each patient can have two possible modalities (A and B), which define a feature space. For instance, modality A could be Chest X-ray imaging data, and modality B could be electronic health records. Each available data modality is associated with a unimodal encoder. If both modalities are present for some patients within the same hospital, then the hospital also has a multimodal model. Note that the arrows indicate the data flow from the hospital to the BlendFL server. However, for the sake of clarity, the updating of the local models based on the global models (from servers to clients) is not depicted in detail but follows the reverse path from the BlendFL server to the hospitals.

The BlendFL framework, as outlined in Algorithm~\ref{alg:BlendFL}, orchestrates the simultaneous training of horizontal and vertical federated learning across multiple clients with heterogeneous data. Note that, to make our framework generic and context-agnostic, we use the term `client' in Algorithm~\ref{alg:BlendFL}, as it is customary in federated learning literature, which equates to `hospital' in our descriptive example. Following Algorithm~\ref{alg:BlendFL}, at each training epoch, the available local datasets at the hospitals are used sequentially to train the models as follows:
\begin{enumerate}
    \item \textbf{Local training with partial data}. Partial data is utilized to train local unimodal models at each hospital (lines 3-8 in Algorithm \ref{alg:BlendFL}). If a hospital has only one modality then it would process that modality.
    
    \item \textbf{Multimodal global model training with fragmented data}. Each hospital holding fragmented data for a subset of patients performs a forward pass to compute intermediate features, which are then sent to the BlendFL server (lines 9-16 in Algorithm \ref{alg:BlendFL}). If a hospital has only one modality, it would therefore execute only the process for that modality. The BlendFL server aligns (pairs) the intermediate features received from the hospitals for the subset of patients with fragmented data and performs a forward and backward pass on the multimodal model (lines 17-19 in Algorithm \ref{alg:BlendFL}). This completes a training epoch for $g^{v}_M$ (hold at the server). The gradients generated during this backward pass are decoupled and sent back to the respective hospitals to complete a full training cycle (lines 20-23 in Algorithm \ref{alg:BlendFL}).
    
    \item \textbf{Local training with paired data}. Hospitals use their paired data (both modalities are present) to train the local multimodal model (lines 24-29 in Algorithm \ref{alg:BlendFL}).
\end{enumerate}

This concludes a local training epoch for all hospitals using all locally available data. Then, the process continues as follows (lines 30-32 in Algorithm \ref{alg:BlendFL}):

\begin{enumerate}
    \item Unimodal and multimodal model parameters from all hospitals are sent to the server for weight aggregation. 
    \item The server aggregates the model parameters using BlendAvg (see Section~\ref{blendavg}) for both unimodal and multimodal models, as defined in Eq.~\ref{aggregation}, Eq.~\ref{aggregation2}, and Eq.~\ref{blended}.
    \item The parameters of the aggregated models (i.e., $g^{blended}_A$, $g^{blended}_b$, $g^{blended}_M$ in Fig.~\ref{fig:architecture}) are distributed to the hospitals, which update their local models, concluding a global training epoch.
\end{enumerate}


This training process repeats until the pre-defined number of training epochs is reached. This iterative process results in one blended global multimodal model and one blended global unimodal model per modality. Through non-restrictive collaborative learning, seamlessly blending vertical and horizontal federated learning, the BlendFL framework leverages all data available at the clients (hospitals), regardless of their available local sample (patients) and feature (modalities) spaces.

\begin{table*}[ht]
\centering
\caption{Performance results of BlendFL, centralized learning, and federated baselines for multimodal and unimodal predictions on the S-MNIST dataset. The best collaborative framework results are shown in bold.}
\resizebox{\linewidth}{!}{\begin{tabular}{l>{\columncolor[RGB]{251,	218, 226}}c>{\columncolor[RGB]{251,	218, 226}}c>{\columncolor[RGB]{205,	246,	236}}c>{\columncolor[RGB]{205,	246,	236}}c>{\columncolor[RGB]{	255,	241,	209}}c>{\columncolor[RGB]{255,	241,	209}}c}
\toprule
\multirow{2}{*}{\textbf{Method}} & \multicolumn{2}{c}{\textbf{Multimodal}} & \multicolumn{2}{c}{\textbf{Audio}} & \multicolumn{2}{c}{\textbf{Image}} \\
       & \textbf{AUROC} & \textbf{AUPRC} & \textbf{AUROC} & \textbf{AUPRC} & \textbf{AUROC} & \textbf{AUPRC} \\
\midrule
Centralized  & 0.989 (0.983, 0.994) & 0.928 (0.917, 0.940) & 0.807 (0.781, 0.833) & 0.409 (0.371, 0.446) & 0.982 (0.974, 0.990) & 0.912 (0.901, 0.923) \\
\midrule
FedAvg~\cite{mcmahan2017communication} & 0.956 (0.943, 0.969) & 0.815 (0.798, 0.832) & 0.716 (0.689, 0.743) & 0.281 (0.241, 0.320) & 0.963 (0.950, 0.976)  & 0.860 (0.846, 0.874) \\
FedMA~\cite{wang2020federated} & 0.857 (0.841, 0.873) & 0.497 (0.456, 0.538) & 0.659 (0.630, 0.688) & 0.198 (0.159, 0.237) & 0.945 (0.932, 0.958) & 0.699 (0.664, 0.734) \\
FedProx~\cite{li2020federated} & 0.953 (0.938, 0.968) & 0.808 (0.778, 0.838) & 0.724 (0.695, 0.753) & 0.288 (0.247, 0.329) & 0.962 (0.951, 0.975) & 0.854 (0.828, 0.880) \\
FedNova~\cite{wang2020tackling} & 0.957 (0.942, 0.972) & 0.824 (0.794, 0.854) & 0.722 (0.693, 0.751) & 0.286 (0.245, 0.327) & 0.963 (0.950, 0.976)  & 0.855 (0.829, 0.881) \\
One-Shot VFL~\cite{sun2023communication} & 0.829 (0.813, 0.845) & 0.474 (0.433, 0.515) & 0.630 (0.601, 0.659) & 0.169 (0.130, 0.208) & 0.916 (0.903, 0.929) & 0.670 (0.635, 0.705) \\
HFCL~\cite{elbir2022hybrid} & 0.936 (0.921, 0.951) & 0.742 (0.712, 0.772) & 0.685 (0.656, 0.714) & 0.239 (0.200, 0.278) & 0.945 (0.932, 0.958) & 0.784 (0.751, 0.817) \\
SplitNN~\cite{vepakomma2018split} & 0.942 (0.928, 0.956) & 0.776 (0.758, 0.794) & 0.718 (0.690, 0.746) & 0.273 (0.234, 0.311) & 0.958 (0.948, 0.968) & 0.827 (0.812, 0.842) \\
\rowcolor[gray]{0.9}BlendFL         & \textbf{0.983 (0.977, 0.989)} & \textbf{0.914 (0.902, 0.926)} & \textbf{0.803 (0.777, 0.829)} & \textbf{0.412 (0.374, 0.450)} & \textbf{0.978 (0.969, 0.987)} & \textbf{0.893 (0.881, 0.905)} \\
\bottomrule
\end{tabular}}
\label{tab:smnist}
\end{table*}

\section{Experiments}
\label{sec:experiments}
In this section, we first evaluate our proposed framework for the clinical setting depicted in Fig.~\ref{fig:architecture} on two multimodal tasks using a widely-used real-world clinical dataset. Next, we evaluate the generalization of our proposal on another dataset and different multimodal architecture. We also conduct convergence experiments for our novel averaging method, BlendAvg, and perform ablation studies to assess the impact of data distribution and number of clients for BlendFL and relevant baselines. For reproducibility, we make our code publicly available at \href{https://github.com/nyuad-cai/BlendFL}{https://github.com/nyuad-cai/BlendFL}.

\subsection{Datasets and Tasks}
\label{sec:datasets}

For the clinical tasks, we use the \textbf{MIMIC-IV} \cite{johnson2023mimic} and \textbf{MIMIC-CXR}  \cite{johnson2019mimic} datasets, which are real-world clinical datasets consisting of Electronic Health Record (EHR) data for over 65,000 patients admitted to an Intensive Care Unit (ICU) and 377,100 Chest X-Ray (CXR) images, respectively. We paired the imaging data with associated clinical time-series data. We follow the same data splits and clinical tasks used by \cite{hayat2022medfuse}, which are described as follows:

\begin{itemize}
    \item \textbf{Clinical conditions prediction:} This multilabel classification task aims to predict a set of 25 different clinical conditions for each ICU stay. The task utilizes time-series data from the entire ICU record paired with the last CXR image collected during the same stay. The output is a vector of 25 binary phenotype labels, indicating the presence of one or more conditions for a given patient. We used a total of 42,636 EHR samples and 124,740 CXR samples. The data was divided into 70\% training, 10\% validation, and 20\% test sets.
    \item \textbf{In-hospital mortality prediction:} This binary classification task predicts in-hospital mortality based on clinical data from the first 48 hours of an ICU stay. Only stays longer than 48 hours are considered, and each instance is paired with the last CXR image collected during the ICU stay. This task utilized 18,843 EHR samples and 124,740 CXR samples, divided using the same split as for the clinical conditions task.
\end{itemize}

To assess the generalization of our approach to other tasks, we used \textbf{S-MNIST} \cite{smnistdataset}, which is an audio-visual dataset designed for benchmarking multimodal classification. It pairs the original MNIST dataset
\cite{lecun1998mnist} with a spoken digits database
 from Google Speech Commands \cite{warden2018speech}. For our experiments, we randomly sample a subset of the original training set (500 instances). A smaller training dataset enabled us to simulate realistic scenarios, providing a more stringent test of each model’s ability to generalize from smaller, less comprehensive datasets. It also improved our analysis of model behavior under constrained data, common in federated learning. The validation and test sets both consist of 10,000 instances.

 \begin{figure}[t]
\centerline{\includegraphics[width=0.8\columnwidth]{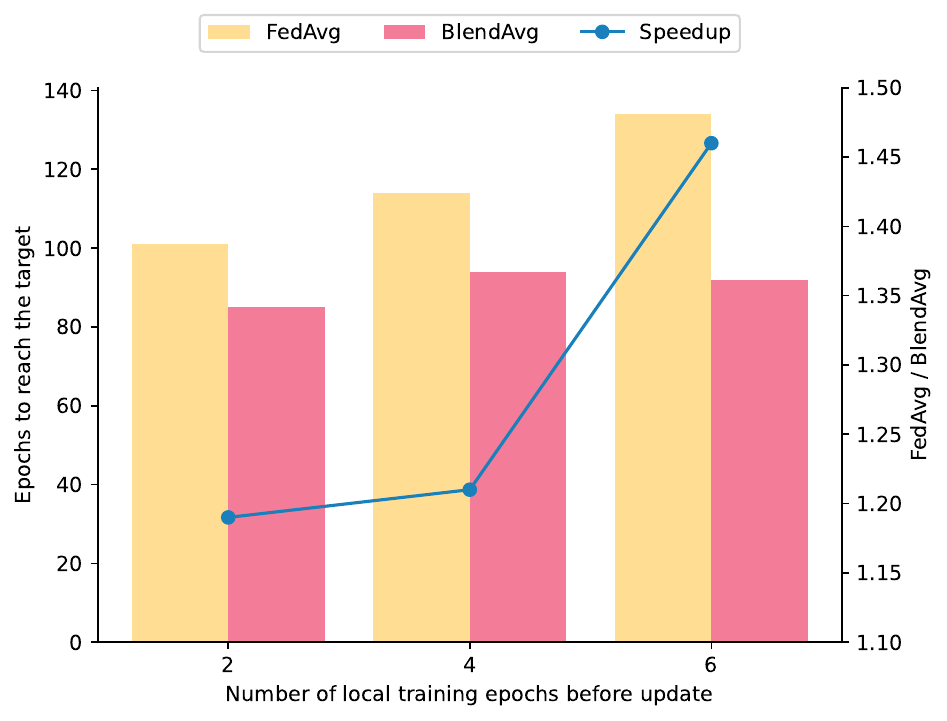}}
\caption{\textbf{Model Convergence}. 
Comparison of rounds needed for model convergence (to reach the target 0.98 AUROC) for the federated model update strategies BlendAvg and FedAvg on the S-MNIST dataset.}
\label{fig:distributions}
\end{figure}

\subsection{Model architectures}

For both clinical tasks using MIMIC-IV and MIMIC-CXR (multilabel and binary classification), we used the architecture proposed by \cite{hayat2022medfuse}, which consists of an LSTM encoder as the EHR data feature extractor and a ResNet-34 \cite{he2016deep} as the CXR image feature extractor. The features are then processed and fused to compute the final multilabel/binary prediction. 

For the multiclass classification task with S-MNIST, we used a multimodal fusion architecture composed of two ResNet-18 \cite{he2016deep} encoders as unimodal (audio and image) feature extractors. The unimodal features are concatenated and used as input for a linear layer that provides the final multiclass prediction. 

\begin{figure*}[ht]
\centerline{\includegraphics[width=0.8\linewidth]{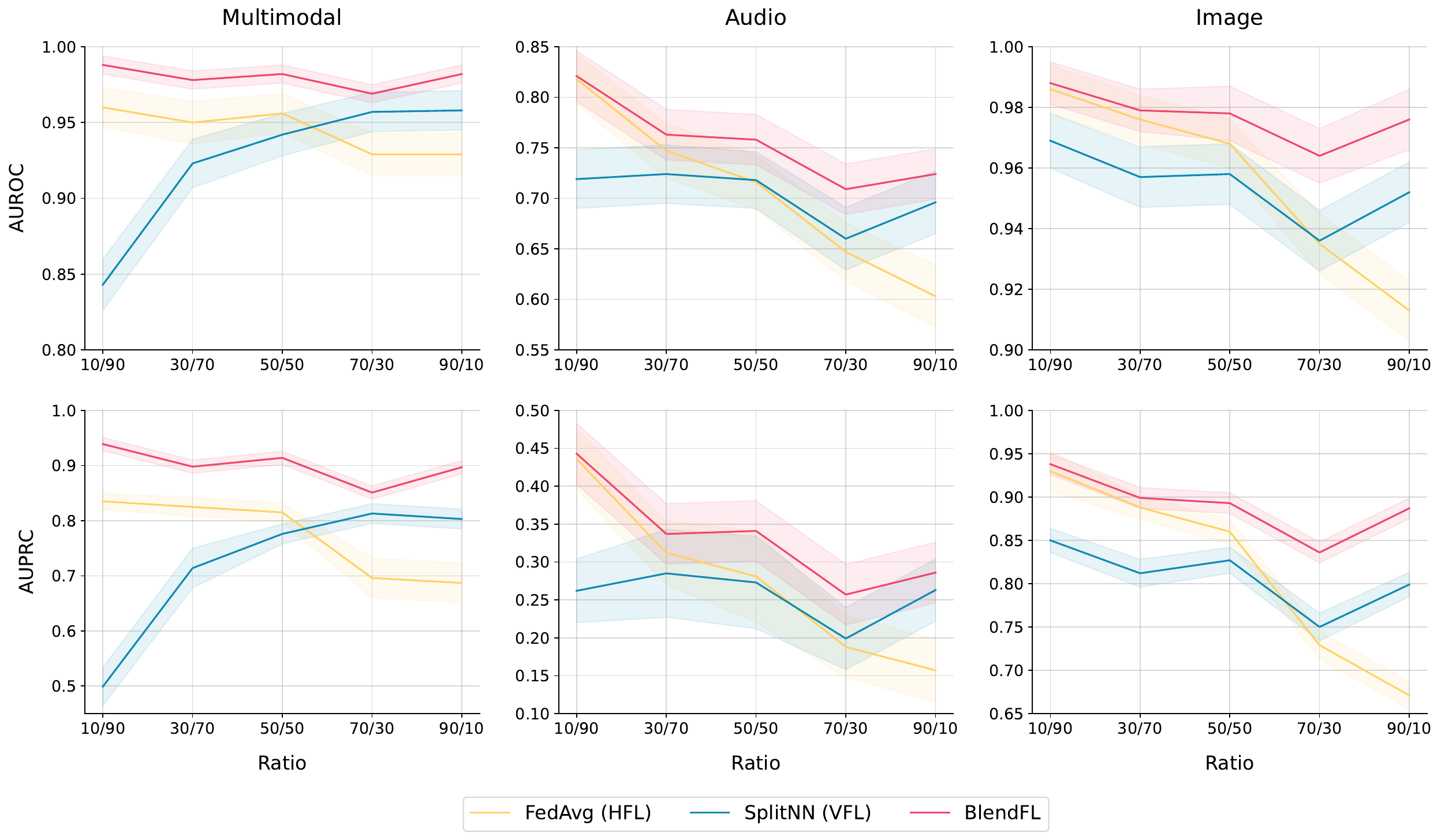}}
\caption{\textbf{Data Distribution}. 
Performance comparison of BlendFL and FL baselines for different data distribution ratios (paired/partial) on the S-MNIST dataset.}
\label{fig:speedup}
\end{figure*}

\subsection{Baselines}

The proposed hybrid FL methodologies focus on improving specific aspects of HFL or VFL frameworks, such as synchronization \cite{gao2023fedhd} or parameter matching \cite{zhang2020hybrid}, by developing additional features for particular characteristics of unimodal data, without considering a holistic integration of HFL and VFL in a multimodal setting as BlendFL does. For that reason, more appropriate baselines for our proposed framework are the foundational FL frameworks such as horizontal federated learning, vertical federated learning, and also centralized learning. The baselines are briefly described as follows:

\begin{itemize}
    \item \textbf{FedAvg~\cite{mcmahan2017communication}} is the foundational form of Federated Learning (FL), where each client trains a local model on its own dataset and periodically shares model updates with a central server. The server aggregates these updates to create a global model, which is sent back to the clients.
    \item \textbf{FedMA~\cite{wang2020federated}} is an optimized form of HFL for CNN and LSTM architectures. It introduces \emph{matched averaging}, which constructs the shared model by matching and aggregating hidden elements in a layer-wise fashion.
    \item \textbf{FedProx~\cite{li2020federated}} is an optimized form of HFL that introduces reparametrization and a proximal term to generalize FedAvg \cite{mcmahan2017communication} to tackle heterogeneity in federated networks.
    \item \textbf{FedNova~\cite{wang2020tackling}} is a form of HFL that combines the FedProx and FedAvg approaches, focusing on addressing objective inconsistency and bias by proposing a novel weighting scheme for local models during averaging.
    \item \textbf{HFCL~\cite{elbir2022hybrid}} is a hybrid FL framework where only clients with enough computational resources are involved in the FL training process. The remaining clients share their data with a central server that performs training on their behalf.
    \item \textbf{SplitNN~\cite{vepakomma2018split}} is the most common implementation of VFL, where parties train a collaborative model in which different clients hold different subsets of features for the same samples
    \item \textbf{One-Shot VFL~\cite{sun2023communication}} is a VFL framework that proposes local semi-supervised learning to address the problems of high communication cost and limited data overlap.
    \item \textbf{Centralized learning} is the traditional ML approach where all data is pooled into a single central server that performs model training. It assumes no privacy concerns and unrestricted data access. Although often infeasible in many real-world applications due to privacy issues, it serves as a strong baseline and performance upper bound for evaluating federated learning methods.
\end{itemize}


\subsection{Convergence and Speedup Experiments}
\label{sec:convergence_experiments}

To evaluate the efficiency and effectiveness of our proposed BlendFL framework, we conducted convergence experiments to compare FedAvg with BlendAvg. These experiments are crucial for understanding the impact of aggregation strategies on the rate of convergence in federated settings. Methods that enable faster convergence reduce communication overhead, lower associated energy consumption, improve scalability, and enhance privacy preservation (due to less time of exposure to possible attacks). In our experimental setup, we measured the number of training rounds (epochs) required to reach a target performance metric, i.e., AUROC of 0.98, using FedAvg and BlendAvg. We quantify the speedup ratio as:

\[
\text{Speedup} = \frac{\text{\# epochs to reach the target using FedAvg}}{\text{\# epochs to reach the target using BlendAvg}}
\]



\subsection{Ablation Study}
\label{sec:ablation_study}

We study the impact of data distributions in terms of imbalance (proportion of paired/partial instances for training) and the number of participating clients on the performance of the BlendFL framework. Specifically, the evaluated scenarios are described as follows:

\begin{itemize}
    \item \textbf{Data distribution}: We evaluated five different ratios of paired-to-partial data distributions: (i) 90/10, (ii) 70/30, (iii) 50/50, (iv) 30/70, and (v) 10/90.
    \item \textbf{Number of clients}: We evaluated the scalability and robustness of BlendFL with a varying number of clients: 4, 8, 12.
\end{itemize}
  

Note that, for efficiency and interpretability, we conducted all these experiments with the S-MNIST dataset and compared BlendFL to the main FL paradigms, HFL and VFL, using implementations of FedAvg \cite{mcmahan2017communication} and SplitNN \cite{vepakomma2018split}, respectively.

\begin{figure*}[ht]
\centerline{\includegraphics[width=0.8\linewidth]{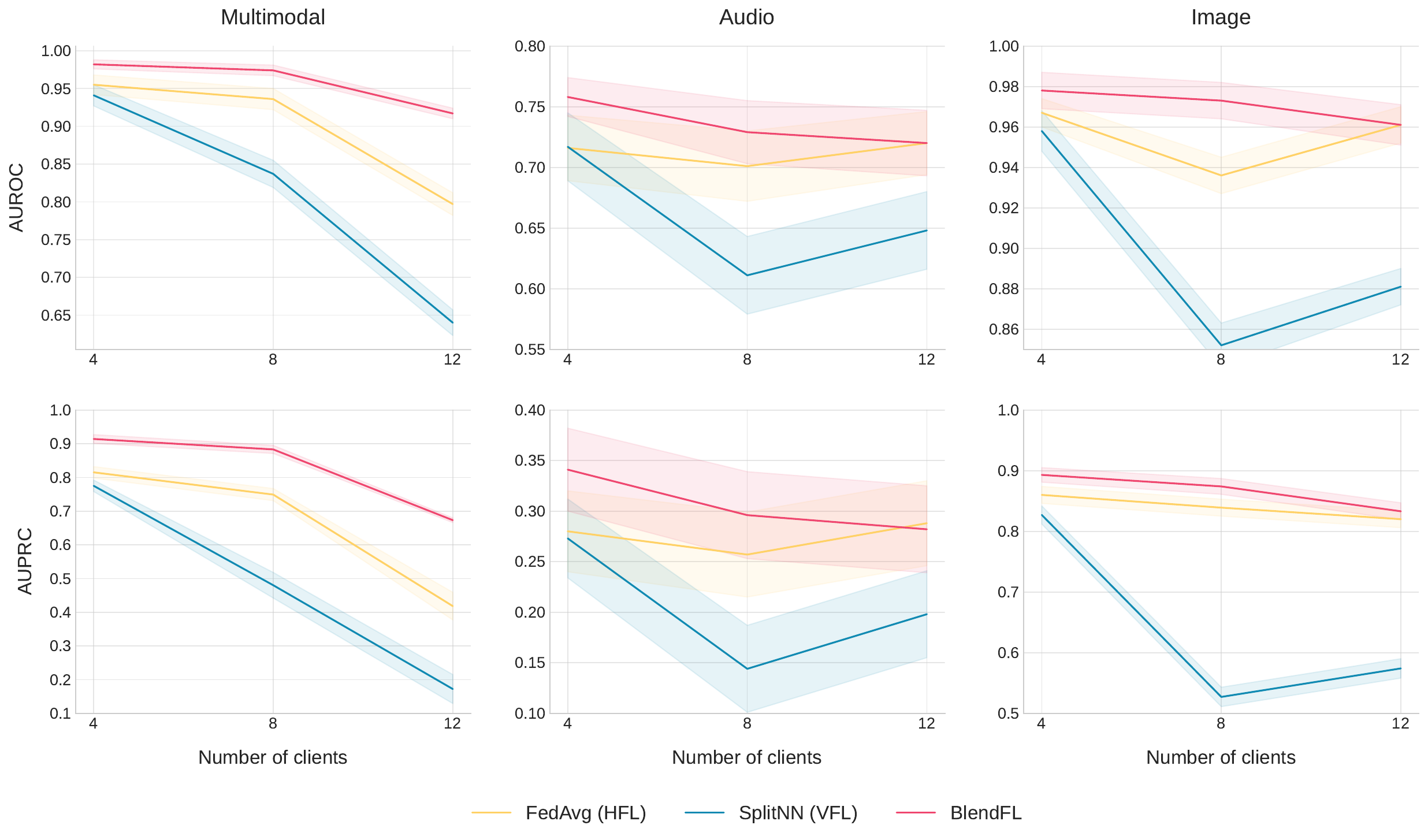}}
\caption{\textbf{Number of clients}. Performance comparison of BlendFL and FL baselines for varying numbers of clients on the S-MNIST dataset.}
\label{fig:nr_nodes}
\end{figure*}

\section{Results}

This section reports the results for the tasks evaluated, including the real-world clinical tasks, generalization to additional dataset, and ablation studies.

\subsection{Real-world medical tasks}

\textbf{Clinical conditions prediction}. Table \ref{tab:pheno} reports the results for clinical conditions prediction on the clinical test set. Overall, BlendFL demonstrates superior performance for collaborative models, consistently outperforming all state-of-the-art FL baseline methods in terms of AUROC and AUPRC metrics. It also exhibits performance close to centralized learning across all metrics (upper bound for FL methods, with all data directly accessible for training). These results underscore the capability of BlendFL to handle complex, multilabel classification tasks effectively, approximating the ideal scenario of complete data availability.



\textbf{In-hospital mortality prediction}. Table \ref{tab:mortality} reports the results for in-hospital mortality prediction on the test set. As in the other task, BlendFL shows superior performance compared to all state-of-the-art FL baselines in terms of AUROC (unimodal and multimodal predictions), indicating its effectiveness in leveraging the full potential of the available heterogeneous data in a binary task. Regarding AUPRC, it outperforms all baselines for unimodal CXR and performs on par with other FL baselines for multimodal and unimodal EHR. Similarly, BlendFL achieves AUROC and AUPRC scores comparable to those of the centralized model.

\subsection{Additional results on S-MNIST}

Table \ref{tab:smnist} reports the performance results for BlendFL and baselines on the S-MNIST test set. As for the clinical tasks, BlendFL demonstrates superior performance for collaborative models, outperforming all state-of-the-art baseline FL methods across all metrics. Note that in this case, the performance gap between the baselines and BlendFL is greater than for the clinical tasks. It also achieves AUROC and AUPRC scores that are on par with those of centralized learning. These results demonstrate the efficacy of BlendFL in contexts where dataset size is constrained and modal variations are significant, highlighting its robustness and adaptability in various multimodal learning scenarios.

\subsection{Model Convergence}

Figure \ref{fig:speedup} reports the results for model updating strategies (BlendAvg and FedAvg) for varying intervals of local training epochs between updates. As the interval increases, the speedup on model convergence gained by using BlendAvg over FedAvg also increases, peaking at a 46\% speedup when updates are made every 6 epochs of local training. This indicates that BlendAvg benefits from allowing local models to train more extensively before averaging, significantly reducing communication overhead while improving model convergence. 


\subsection{Ablation Studies}


\textbf{Data distribution}. Figure~\ref{fig:distributions} shows the impact of different data distributions in terms of paired/partial data on the performance of BlendFL and other baselines. Splits with a higher proportion of paired data favor VFL (SplitNN), reflecting their reliance on comprehensive feature sets per sample. Splits that favor partial data enhance HFL (FedAvg) performance, capitalizing on its strength in leveraging larger volumes of data for the same feature set. Additionally, BlendFL outperforms the baselines in each setting, effectively addressing the weaknesses of each one and maximizing model performance regardless of the distribution of the data.

\textbf{Number of clients}. Figure \ref{fig:nr_nodes} reports the performance of BlendFL and the baselines when varying the number of clients. HFL methods generally perform better relative to VFL approaches as the number of clients increases, benefiting from the aggregation of more sample-diverse datasets. VFL approaches tend to underperform relative to HFL in scenarios with more clients due to the complexity of managing more extensive feature sets across common samples.

\section{Discussion and Future Work}

Training collaborative models in real-world environments is a challenging task. While HFL and VFL enable training under certain conditions imposed on the clients, they fail to provide a framework for ill-defined scenarios where neither all features nor all samples are uniformly available across clients. To address this gap, we introduce BlendFL, the first FL framework that, unlike hybrid FL approaches, seamlessly integrates the strengths and full capabilities of both HFL and VFL for multimodal collaborative training. BlendFL enables collaborative training for diverse clients, allowing them to benefit from HFL, VFL, or both, regardless of their share of features and samples, and without restrictive requirements. Unlike VFL, it also enables local, independent inference, reducing dependency on the server for inference purposes and minimizing communication overhead.

We evaluate the performance of BlendFL using two datasets and three tasks. Our results demonstrate that BlendFL is superior to state-of-the-art FL frameworks under heterogeneous data conditions, consistently outperforming all state-of-the-art baselines across various datasets and tasks. In addition, BlendFL produces multimodal and unimodal encoders with performance on par with centralized models, considered the upper bound for FL. A key factor behind BlendFL's success is BlendAvg, a novel model parameter averaging strategy that shows faster convergence than FedAvg. The ablation studies, which evaluate BlendFL and baseline methods under varying data distributions (ratio of paired/partial data) and client numbers, highlight BlendFL's superior performance across various imbalance data and challenging conditions. Despite BlendFL’s superior performance over all FL baselines, it is not immune to privacy threats. Future work should focus on integrating additional privacy measures into BlendFL, such as differential privacy, to strengthen data privacy and tighten security constraints within the framework.

\section{Acknowledgment}

This work was supported by the NYUAD Center for Interacting Urban Networks (CITIES), funded by Tamkeen under the NYUAD Research Institute Award CG001, the Center for Cyber Security (CCS), funded by Tamkeen under NYUAD RRC Grant No. G1104, and the NYUAD Center for Artificial Intelligence and Robotics, funded by Tamkeen under the NYUAD Research Institute Award CG010.  The research was carried out on the High Performance Computing resources at New York University Abu Dhabi.

This work was also supported by the Mubadala-NYUAD Collaborative Student Research Project Award.

\bibliographystyle{IEEEtran}
\bibliography{ijcnn/bibliography}

\begin{thebibliography}{10}
\providecommand{\url}[1]{#1}
\csname url@samestyle\endcsname
\providecommand{\newblock}{\relax}
\providecommand{\bibinfo}[2]{#2}
\providecommand{\BIBentrySTDinterwordspacing}{\spaceskip=0pt\relax}
\providecommand{\BIBentryALTinterwordstretchfactor}{4}
\providecommand{\BIBentryALTinterwordspacing}{\spaceskip=\fontdimen2\font plus
\BIBentryALTinterwordstretchfactor\fontdimen3\font minus \fontdimen4\font\relax}
\providecommand{\BIBforeignlanguage}[2]{{%
\expandafter\ifx\csname l@#1\endcsname\relax
\typeout{** WARNING: IEEEtran.bst: No hyphenation pattern has been}%
\typeout{** loaded for the language `#1'. Using the pattern for}%
\typeout{** the default language instead.}%
\else
\language=\csname l@#1\endcsname
\fi
#2}}
\providecommand{\BIBdecl}{\relax}
\BIBdecl

\bibitem{yue2020deep}
L.~Yue, D.~Tian, W.~Chen, X.~Han, and M.~Yin, ``Deep learning for heterogeneous medical data analysis,'' \emph{World Wide Web}, vol.~23, pp. 2715--2737, 2020.

\bibitem{milasheuski2024impact}
U.~Milasheuski, L.~Barbieri, B.~C. Tedeschini, M.~Nicoli, and S.~Savazzi, ``On the impact of data heterogeneity in federated learning environments with application to healthcare networks,'' in \emph{2024 IEEE Conference on Artificial Intelligence (CAI)}.\hskip 1em plus 0.5em minus 0.4em\relax IEEE, 2024, pp. 1017--1023.

\bibitem{esteva2019guide}
A.~Esteva, A.~Robicquet, B.~Ramsundar, V.~Kuleshov, M.~DePristo, K.~Chou, C.~Cui, G.~Corrado, S.~Thrun, and J.~Dean, ``A guide to deep learning in healthcare,'' \emph{Nature medicine}, vol.~25, no.~1, pp. 24--29, 2019.

\bibitem{rieke2020future}
N.~Rieke, J.~Hancox, W.~Li, F.~Milletari, H.~R. Roth, S.~Albarqouni, S.~Bakas \emph{et~al.}, ``The future of digital health with federated learning,'' \emph{NPJ digital medicine}, vol.~3, no.~1, pp. 1--7, 2020.

\bibitem{wen2023survey}
J.~Wen, Z.~Zhang, Y.~Lan, Z.~Cui, J.~Cai, and W.~Zhang, ``A survey on federated learning: challenges and applications,'' \emph{International Journal of Machine Learning and Cybernetics}, vol.~14, no.~2, pp. 513--535, 2023.

\bibitem{mcmahan2017communication}
B.~McMahan, E.~Moore, D.~Ramage, S.~Hampson, and B.~A. y~Arcas, ``Communication-efficient learning of deep networks from decentralized data,'' in \emph{Artificial intelligence and statistics}.\hskip 1em plus 0.5em minus 0.4em\relax PMLR, 2017.

\bibitem{dayan2021federated}
I.~Dayan, H.~R. Roth, A.~Zhong, A.~Harouni, A.~Gentili \emph{et~al.}, ``Federated learning for predicting clinical outcomes in patients with covid-19,'' \emph{Nature medicine}, vol.~27, no.~10, pp. 1735--1743, 2021.

\bibitem{chen2020vafl}
T.~Chen, X.~Jin, Y.~Sun, and W.~Yin, ``Vafl: a method of vertical asynchronous federated learning,'' \emph{arXiv:2007.06081}, 2020.

\bibitem{yang2019federated}
Q.~Yang, Y.~Liu, T.~Chen, and Y.~Tong, ``Federated machine learning: Concept and applications,'' \emph{ACM Trans. on Intel. Sys. and Tech.}, 2019.

\bibitem{banabilah2022federated}
S.~Banabilah, M.~Aloqaily, E.~Alsayed, N.~Malik, and Y.~Jararweh, ``Federated learning review: Fundamentals, enabling technologies, and future applications,'' \emph{Information processing \& management}, no.~6, 2022.

\bibitem{karimireddy2020scaffold}
S.~P. Karimireddy, S.~Kale, M.~Mohri, S.~Reddi, S.~Stich, and A.~T. Suresh, ``Scaffold: Stochastic controlled averaging for federated learning,'' in \emph{International conference on machine learning}.\hskip 1em plus 0.5em minus 0.4em\relax PMLR, 2020.

\bibitem{wang2020federated}
H.~Wang, M.~Yurochkin, Y.~Sun, D.~Papailiopoulos, and Y.~Khazaeni, ``Federated learning with matched averaging,'' \emph{arXiv:2002.06440}, 2020.

\bibitem{zhang2020hybrid}
X.~Zhang, W.~Yin, M.~Hong, and T.~Chen, ``Hybrid federated learning: Algorithms and implementation,'' \emph{arXiv:2012.12420}, 2020.

\bibitem{li2023effective}
Q.~Li, C.~Xie, X.~Xu, X.~Liu, C.~Zhang \emph{et~al.}, ``Effective and efficient federated tree learning on hybrid data,'' \emph{arXiv:2310.11865}, 2023.

\bibitem{hu2019fdml}
Y.~Hu, D.~Niu, J.~Yang, and S.~Zhou, ``Fdml: A collaborative machine learning framework for distributed features,'' in \emph{Proceedings of the 25th ACM SIGKDD Intl. Conf. on Knowl. Disc. \& Data Mining}, 2019.

\bibitem{vepakomma2018split}
P.~Vepakomma, O.~Gupta, T.~Swedish, and R.~Raskar, ``Split learning for health: Distributed deep learning without sharing raw patient data,'' \emph{arXiv:1812.00564}, 2018.

\bibitem{gao2023fedhd}
H.~Gao, S.~Ge, and T.-H. Chang, ``Fedhd: Communication-efficient federated learning from hybrid data,'' \emph{Journal of the Franklin Institute}, vol. 360, no.~12, pp. 8416--8454, 2023.

\bibitem{zhang2023privacy}
H.~Zhang, J.~Hong \emph{et~al.}, ``A privacy-preserving hybrid federated learning framework for financial crime detection,'' \emph{arXiv:2302.03654}, 2023.

\bibitem{morales2023private}
D.~Morales, I.~Agudo, and J.~Lopez, ``Private set intersection: A systematic literature review,'' \emph{Computer Science Review}, vol.~49, 2023.

\bibitem{li2020federated}
T.~Li, A.~K. Sahu, M.~Zaheer, M.~Sanjabi, A.~Talwalkar, and V.~Smith, ``Federated optimization in heterogeneous networks,'' \emph{Proceedings of Machine learning and systems}, vol.~2, pp. 429--450, 2020.

\bibitem{wang2020tackling}
J.~Wang, Q.~Liu, H.~Liang, G.~Joshi, and H.~V. Poor, ``Tackling the objective inconsistency problem in heterogeneous federated optimization,'' \emph{Advances in neural information processing systems}, vol.~33, 2020.

\bibitem{sun2023communication}
J.~Sun, Z.~Xu, D.~Yang, V.~Nath, W.~Li, C.~Zhao, D.~Xu, Y.~Chen, and H.~R. Roth, ``Communication-efficient vertical federated learning with limited overlapping samples,'' in \emph{Proceedings of the IEEE/CVF International Conference on Computer Vision}, 2023, pp. 5203--5212.

\bibitem{elbir2022hybrid}
A.~M. Elbir, S.~Coleri, A.~K. Papazafeiropoulos, P.~Kourtessis, and S.~Chatzinotas, ``A hybrid architecture for federated and centralized learning,'' \emph{IEEE Trans. on Cogn. Comm. and Net.}, vol.~8, no.~3, 2022.

\bibitem{johnson2023mimic}
A.~E. Johnson, L.~Bulgarelli, L.~Shen \emph{et~al.}, ``Mimic-iv, a freely accessible electronic health record dataset,'' \emph{Scientific data}, vol.~10, 2023.

\bibitem{johnson2019mimic}
A.~E. Johnson, T.~J. Pollard, S.~J. Berkowitz, N.~R. Greenbaum \emph{et~al.}, ``Mimic-cxr, a de-identified publicly available database of chest radiographs with free-text reports,'' \emph{Scientific data}, vol.~6, p. 317, 2019.

\bibitem{hayat2022medfuse}
N.~Hayat, K.~J. Geras, and F.~E. Shamout, ``Medfuse: Multi-modal fusion with clinical time-series data and chest x-ray images,'' in \emph{Machine Learning for Healthcare Conference}.\hskip 1em plus 0.5em minus 0.4em\relax PMLR, 2022.

\bibitem{smnistdataset}
L.~Khacef, L.~Rodriguez, and B.~Miramond, ``Written and spoken digits database for multimodal learning,'' https://zenodo.org/doi/10.5281/zenodo.3515934, 2019.

\bibitem{lecun1998mnist}
Y.~LeCun, ``The mnist database of handwritten digits,'' \emph{http://yann. lecun. com/exdb/mnist/}, 1998.

\bibitem{warden2018speech}
P.~Warden, ``Speech commands: A dataset for limited-vocabulary speech recognition,'' \emph{arXiv:1804.03209}, 2018.

\bibitem{he2016deep}
K.~He, X.~Zhang, S.~Ren, and J.~Sun, ``Deep residual learning for image recognition,'' in \emph{Proceedings of the IEEE Conf. on Comp. Vision and Pattern Recognition}, 2016, pp. 770--778.

\end{thebibliography}

\end{document}